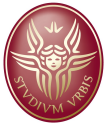
# SAPIENZA
## Università di Roma

# A pilot study on the daily control capability of s-EMG prosthetic hands by amputees

**Facoltà di Ingegneria dell'Informazione, Informatica e Statistica**
**Corso di laurea in Ingegneria dell'Informazione**
Cattedra di Informatica Teorica
Francesca Giordaniello
1541253

Relatore
Barbara Caputo

Correlatore
Umberto Nanni

A/A 2014/2015

# Index

# Abstract

Surface electromyography is a valid tool to gather muscular contraction signals from intact and amputated subjects. Electromyographic signals can be used to control prosthetic devices in a noninvasive way distinguishing the movements performed by the particular EMG electrodes activity. According to the literature, several algorithms have been used to control prosthetic hands through s-EMG signals. The main issue is to correctly classify the signals acquired as the movement actually performed. This work presents a study on the Support Vector Machine's performance in a short-time period, gained using two different feature representation (Mean Absolute Value and Waveform Length) of the sEMG signals. In particular, we paid close attention to the repeatability problem, that is the capability to achieve a stable and satisfactory level of accuracy in repeated experiments. Results on a limited setting are encouraging, as they show an average accuracy above 73% even in the worst case scenario.

# 1.   Introduction

Human hands are a complex system: they allow us to make a large number of movements with a small effort required. As opposed to this, although an artificial limb may present several degrees of freedom (DoFs) to emulate the dexterity of natural hands, controlling them at a satisfactory level is yet not feasible [1]. In fact, signals classification into movement classes is still being studied to improve prosthetic hands reliability, and scientists and engineers are still studying how to replicate the sensory-motor function of the human hand.

The body's actions are controlled by neural signals sent from the central nervous system (CNS) to the peripheral nervous system (PNS). Muscles are made up of fibers organised in motor units (MUs); each of them includes the single motoneuron that innervates that unit and the MU itself, as illustrated in Figure 1. The more precise the movements produced by the MU are, the lower the number of muscle fibers innervated by the motor neuron is. For example, visual fixation of objects by the ocular muscles will involve only a few muscle fibers being supplied by a motor neuron, while movements produced by limb muscles will have a motor neuron innervating many muscle fibers.

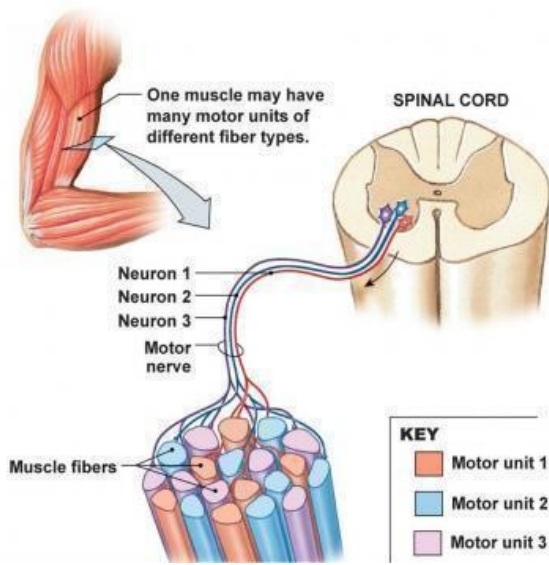

Figure 1. The arrangement of motor units in a skeletal muscle.

The electrical impulse activates a motor unit and causes graded muscle contraction related to the number of MUs involved.
Electromyography (EMG) is a noninvasive technique to measure the electrical potentials of the MUs; signals are gathered by electrodes placed over the skin, above appropriate muscles.

Currently transradial-amputation subjects can rely on prostheses controlled by surface electromyographic signals (sEMG); they have been used since 1948 and they're still the most common approach to control upper limb prosthetic devices because they provide a simple and noninvasive access to the processes that cause muscular contraction [2].

Nowadays prostheses ensure a multifunctional control instead of a single function (such as "open-close" grasp). Nevertheless it has been found a lack of acceptance among amputees due to an insufficient level of dexterity in daily-life, a lack of sensory feedback and the extreme effort required to carry out even the simplest movements [3]. To improve functionality it is necessary to focus on extracting relevant information from the myoelectric signals and their classification in movement classes, as stated in [3].

In this study we specifically examined the multi-class classification problem, in particular considering the repeatability of the experiments during a short-time period. The present work contains a brief overview of related works up to now and a clarification of the problem analyzed. Then the dataset used is depicted and the features and classifier employed are presented, followed by the description of the experiments and the discussion of what has been achieved. Further graphs and tables are given in the Appendix to show the results.

## 2. Related work

There is a large amount of work related to sEMG controlled devices that focus on the classification of muscular contractions for myoelectric control of the prostheses. In all these approaches it was followed a common process that includes data acquisition and preprocessing, feature extraction and finally movement classification [1, 3, 4, 5, 11].

Recently it has been built up the Ninapro database, in which are collected data acquired from 27 healthy subjects and referred to 52 different movements, including fingers and wrist movements, hand postures and grasping gestures [4, 11]. This dataset leads to several advantages since it enhances the number of both movements and subjects compared to previous studies. Moreover, it's a public database, therefore it concurs to the exchange of knowledge, information and results gathered using various methods.

The Ninapro dataset was acquired using ten sEMG electrodes; eight of them were placed beneath the elbow at a fixed distance from the radio-humeral joint, while the other two were placed on the flexor and extensor muscles.

In addition, a two-axes inclinometer and a Cyberglove were used to gather more information over the potential generated by muscular contraction. Figure 2 presents the acquisition set placement.

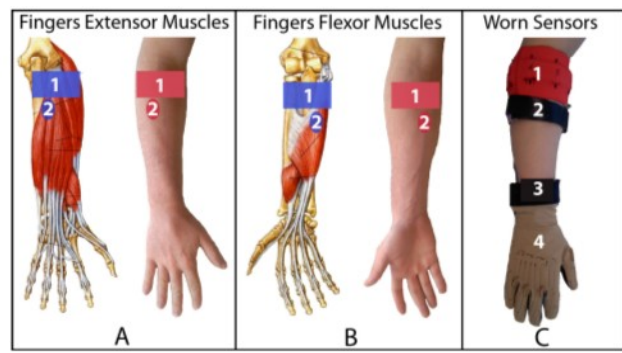

Figure 2. Setting of the electrodes (1, 2) on fingers extensor (A) and flexor (B) muscles. Electrodes (1, 2), Inclinometer (3) and Cyberglove (4) positioned on the arm are showed in (C). Figure taken from [4].

The information is extracted from the signal and represented in a particular feature, so that relevant data density of the EMG signals is increased, noise and irrelevant data are discarded and the output classes are better discernible.

The extraction methods employed in previous works operate in time, frequency or both time-frequency domain.

Examples of time domain operating methods are Mean Absolute Value, Variance, Waveform Length and Cepstral Coefficient. They are generally quickly calculated since they don't imply a transformation. Frequency domain is studied in methods such as Frequency Ratio and Mean Frequency. The time-frequency features are more sophisticated and computationally expensive. They include Short Time Fourier Transform, Wavelet Transform and Wavelet Packet Transform. We refer the reader to [2] and [6] for further details about the feature extraction methods mentioned.

There are several possible classifiers. Between all of them, the most popular in the literature are: Linear Discriminant Analysis, k-Nearest Neighbours, Multi-Layer Perceptron and Support Vector Machine [3].

All these methods have been studied in previous works, combining several feature extractors with classificators of different kind. The classification methods examinated were observed to have similar performance [1, 3]. Therefore, the choice among them is mainly led by methods' popularity in literature rather than their computational cost. Feature representation of the data has a greater influence than classifiers complexity because of its capability to capture certain signals information [1].

# 3. Problem statement

To the best of our knowledge there is no prior work investigating the repeatability problem. The term "repeatability" indicates the variation in repeated measurements made on the same subject and under the same conditions. Thus, in a repeatability study, the variability is only due to errors in the measurement process itself, as stated in [9].

The aim of this work is to contribute studying this issue, particularly focusing on the performance of the SVM algorithm in classifying 18 movements (including rest) during a short-time period. Indeed, managing to distinguish each movement type with high accuracy in variable conditions implies being able to built a reliable limb prostheses.

Making the sEMG signals capable to control a prosthetic device requires several steps of processing. Data acquired are preprocessed in order to remove noise due to the electrodes and the electronic equipment. In particular, filtering the signals gives "clearer" data to be processed.
The windowing phase guarantees removing samples with an ambiguous label, that is those recorded during the transition from rest to the actual movement. Right after the feature extraction, the signal is represented highlighting one of its particular property. Choosing one feature in preference to the others should be motivated by better chances to discriminate different movement classes based on that characteristic.
The final step consists of classifying the movement labels using a pattern recognition algorithm. Since no relevant difference has been found between the well-known classifiers, we selected Support Vector Machine, which is widely employed because of the possibility to be used joined with kernel functions [3].
The whole process is summarized in Figure 3.

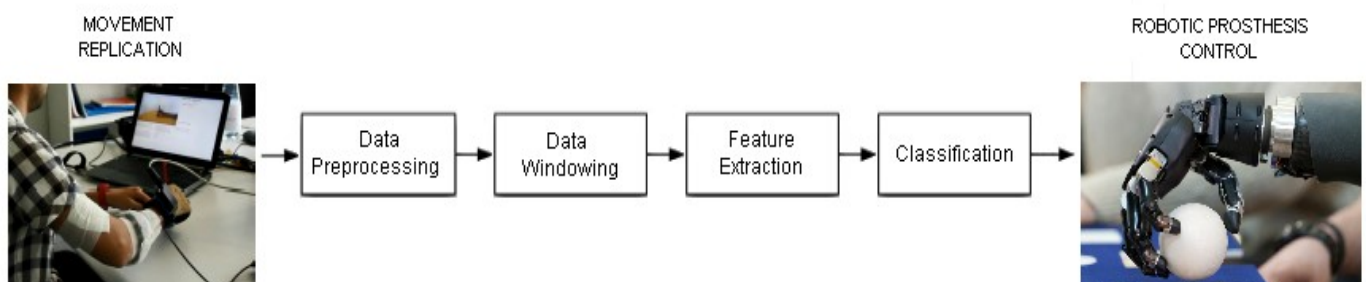

Figure 3. Scheme giving a guideline of the steps followed for data processing. Figures taken from [1], [10].

# 4. Material and Methods

Within the features presented in chapter 2, only two were exploited to represent data in this study. Furthermore, a single method was chosen to perform data classification.

## 4.1. Data

The data used in this study do not belong to the Ninapro dataset; they were acquired from a single intact subject by sEMG electrodes, cyberglove and inclinometers. The latter two were not examined because of issues with glove's sensors during acquisition (signals were unusable) and due to the different process needed to study the data gathered from inclinometers.

The acquisition period lasted four days, with three daily sessions of data collection. A short movie was presented to the subject, showing the movements to replicate (five seconds each). Only 17 of the 52 Ninapro database postures were considered in this study, specifically those in the Exercise 2. Figure 4 shows stills from the movie, illustrating the movements the subject was asked to replicate.

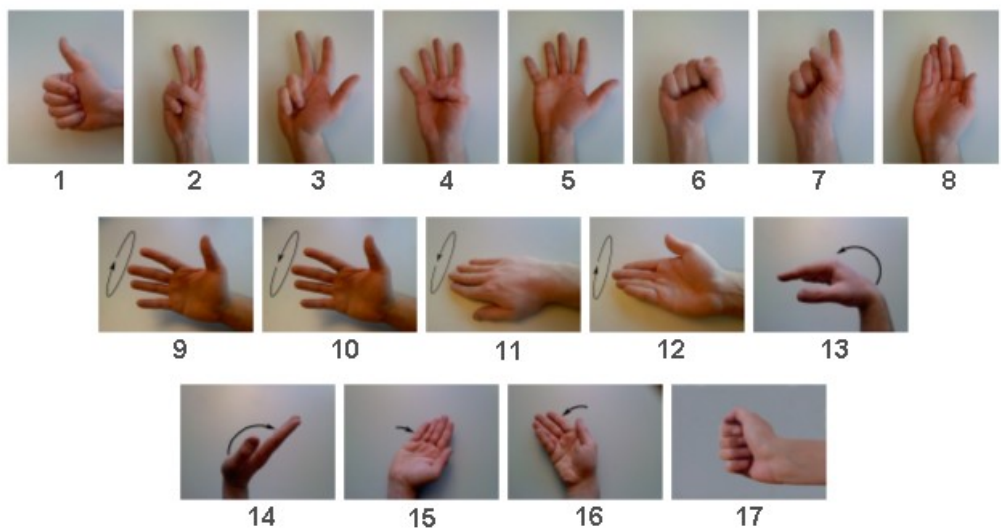

Figure 4. Hand postures and wrist movements performed by the subject during the acquisition phase. Figure taken from [4].

Each hand movement was repeated ten times, with a three seconds rest between the repetitions. Data was recorded with a 100Hz frequency from the electrodes, which provide an initial filtering and amplification of the signal and its Root Mean Square (RMS) version.

Data were stored into three (only two of them used) files in ASCII format. Each row contains a timestamp followed by different kind of values. The first file contains the sEMG sensor values (10 channels), 2 columns for the inclinometer signal and 4 columns for the glove's one (these last 6 data columns were not employed). Movement labels are included in the second file corresponding to each timestamp. The third file gathers more information from the cyberglove and it was not considered in the study as well.

## 4.2. Feature extraction

Two different features were employed for data representation, chosen among the most common in the literature: Mean Absolute Value and Waveform Length.
A portion of the signal (Figure 7) is showed as it appears using MAV and WL in Figure 5 and Figure 6 respectively.

- *Mean Absolute Value*

The Mean Absolute Value (MAV) is a time domain feature that estimates the mean absolute value of the signal $x_i$ in the i-th segment that is N samples in length:

$$X_i = \frac{1}{N} \sum_{k=1}^{N} |x_k| \quad \text{for } i = 1, \ldots, I$$

where $x_k$ is the k-th sample in segment *i* and *I* is the total number of segments over the entire sampled signal. In other words, it is calculated by taking the average of the absolute value of sEMG signal. It is a way to detect muscle contraction levels. MAV does not require a high computational cost so it's widely used in research and in clinical practice [2].

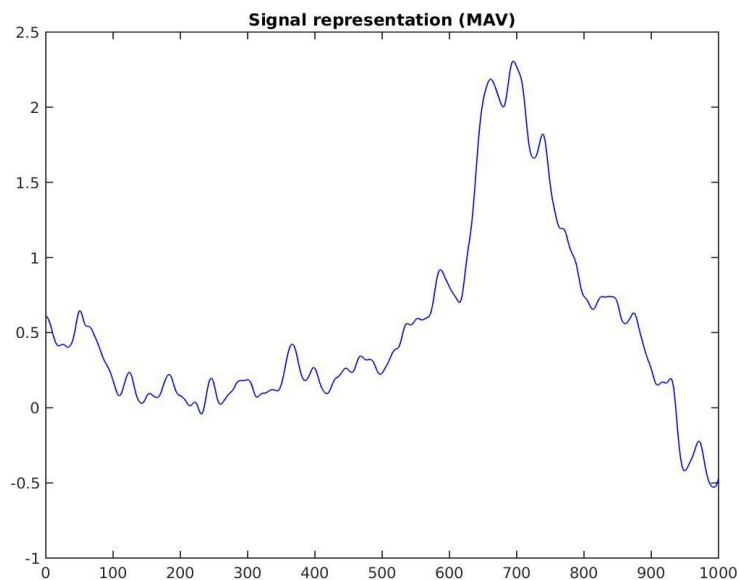

Figure 5. Portion of the signal acquired through one of the electrodes, as it appears after the feature extraction phase using MAV.

- *Waveform Length*

The Waveform Length (WL) provides information on the waveform complexity in each segment in the waveform length. That corresponds to the cumulative length of the waveform over the time segment:

$$l_0 = \sum_{k=1}^{N} |x_k - x_{k-1}|$$

It gives a measure of waveform amplitude, frequency and duration in a single parameter.

It has been used successfully by [6] to classify hand movements from sEMG signals, in which it seemed to be the best feature concerning accuracy rate and stability to different preprocessing.

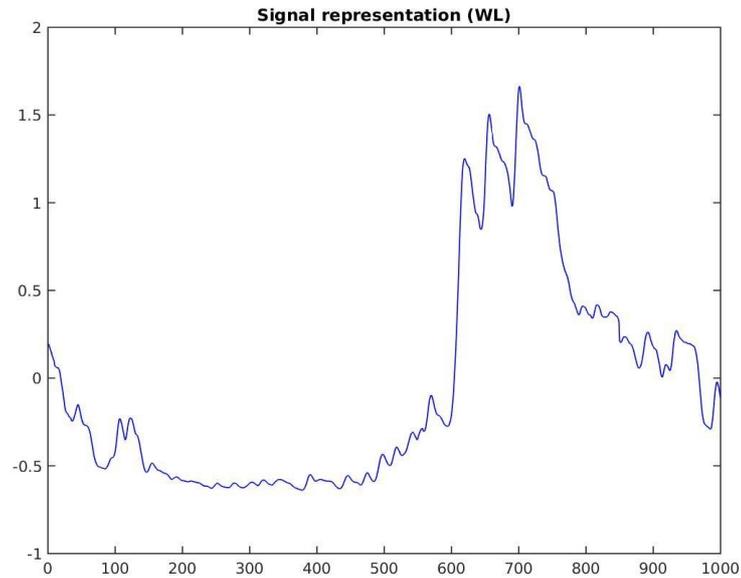

Figure 6. Portion of the signal acquired through one of the electrodes, using WL as feature extractor.

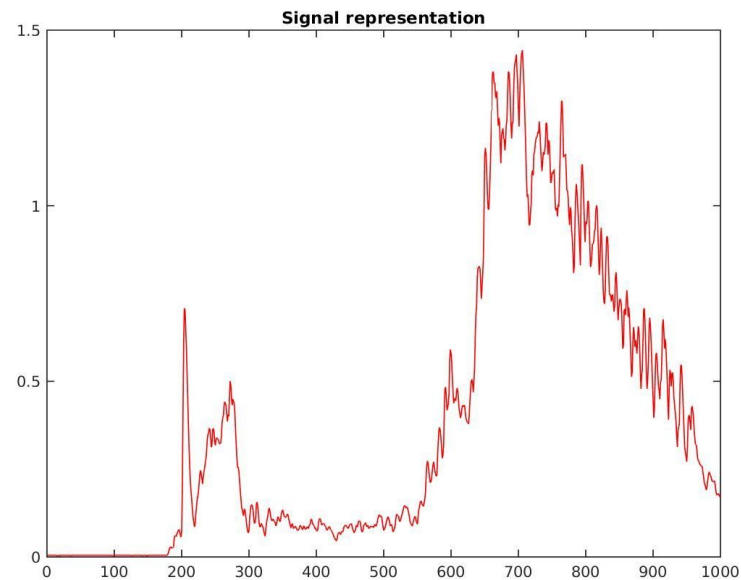

Figure 7. Part of the signal gathered by one of the electrodes before the feature extraction.

### 4.3. Classification

Several studies on sEMG demonstrate that no classification method considerably overcomes the other in pattern recognition, so that linear classifiers' performance is practically equal to the nonlinear [1].
Support Vector Machine (SVM) was selected in this work within the classification approaches known in the literature.

### 4.3.1. Algorithm description

SVMs are linear classifiers that try to maximize the margin between two classes. Although they are binary classifiers, they can also be used in one-vs-one multiclass classification.

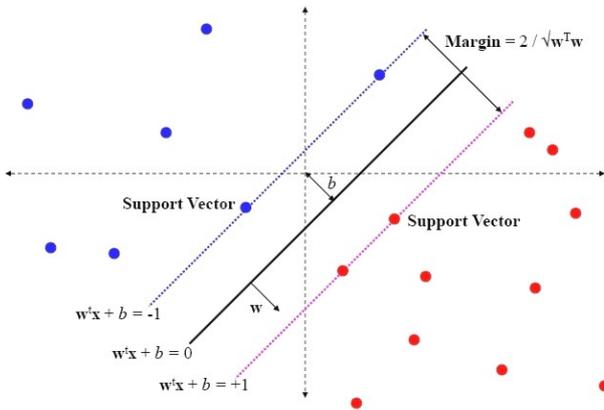

In linear classification the linear machine is given a training dataset associated with labels. The aim of SVM is to produce a model based on the training data that predicts the testing dataset target values, given only the test data attributes. In other words, it finds a hyperplane that maximizes the margin between the inputs of the two distinct classes (Figure 8).

Figure 8. Two classes of linearly separable data. The so-called decision boundary ( $w^T x + b = 0$ ), plus-plane ( $w^T x + b = 1$ ) and minus-plane ( $w^T x + b = -1$ ) are shown. Support vectors are the data points lying on either the minus-plan or the plus-plane.

In spite of their linear approach, SVMs can be used jointly with kernel functions so that they are usually employed in nonlinear problems. In practice, kernel functions perform a transformation, mapping input data into a different space, so that it is easier for SVM to process them and to find an hyperplane separating the two classes data samples. A simple example can be seen in Figure 9.

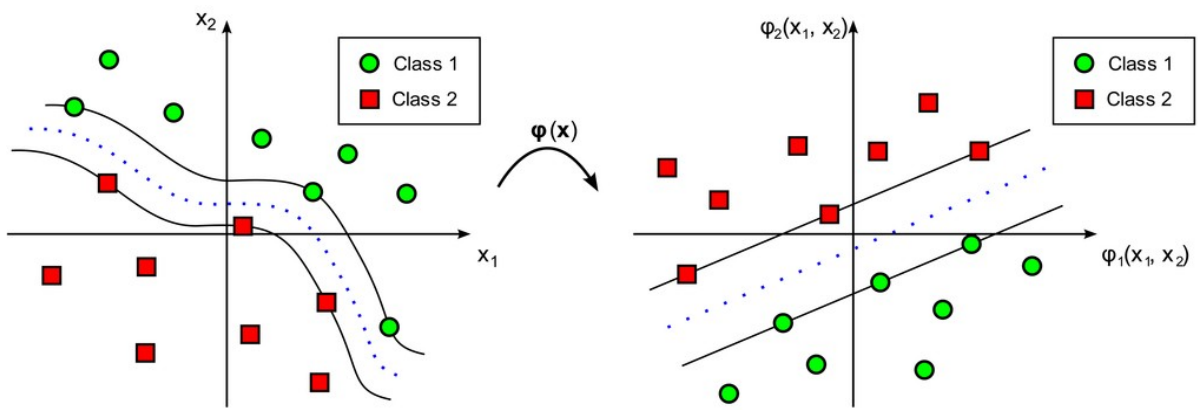

Figure 9. Example of kernel application to an SVM classifier. Function $\varphi$ is the nonlinear transformation mapping vectors from the input space to the feature space.

Radial Basis Function (RBF) kernel was used in this study. It maps data into a higher dimensional space so it can even handle the case of nonlinear relation between class labels.

### 4.3.2. Parameters description

RBF only requests two parameters (called hyperparameters) to be set by grid search: $C$ and $\gamma$. The $C$ parameter controls the trade-off between model complexity and error frequency; in other words, a lower $C$ provides a better margin from the hyperplane (though more samples would fail to be classified correctly), while an increased value aims at separating as many instances as possible (margin may not be the best possible). Intuitively, the γ parameter defines how far the influence of a single training example reaches: the higher are the values, the closer is the distance.

The first value was chosen among $C \in \{2^i : i \in \{0, 2, ..., 14, 16\}\}$, while the latter $\gamma \in \{2^i : i \in \{-16, -14, ..., -4, -2\}\}$. The goal is to choose the pair ($C$, $\gamma$) so that the classifier can accurately predict the testing dataset. For that purpose, the validation set (defined in chapter 5.1) is tested for each value of both the hyperparameters, then $C$ and $\gamma$ are chosen in order to get the best accuracy.

LibSVM [13] has been used during classification since it provides functions to easily implement training, testing and grid search.

# 5. Experiments

Following the acquisition, data was subjected to several steps before the classification phase.

## 5.1. Experimental setup

Although sEMG signals had been pre-filtered during acquisition, we needed to preprocess the data in order to extract the relevant informations and reduce the electrical noise, which is due to different factors (electronic equipment components, signal instability, etc).

First of all, data from the movies and the electrodes needed to be synchronized. For this purpose, linear interpolation between the timestamps was performed.

It was necessary a relabeling process as well. In fact, the movements performed by the subject might not match perfectly with the movements showed in the video; they may begin slightly before the actual start of the movie or have a small delay. This phase ensures the realignment between movement labels and timestamps corresponding to an increase in sEMG activity.
Thereafter, we performed whitening on the signals to remove autoregressive components, which are irrelevant [3]. For this purpose we used a multivariate VAR(p) (Vector Autoregressive) model, with value p = 20, as suggested in [3]. An autoregressive process is a stochastic process which operates assuming that past values affect current values as well. The variable p means that the current value is based on all the previous p values.

After filtering, relevant information lies in the low frequency spectrum. Hence, high-frequency noise component was removed applying a second order low-pass filter, in particular a Butterworth filter at 5Hz, that is commonly used in the literature [3, 4, 5].

Afterward, for each channel, the signal was divided into 100ms length windows, using a single sliding window. In that way, we attempted to avoid transition states to be involved in signal processing.
Selecting the window length implies a trade-off. As stated in [1], a longer window guarantees stability of the feature used, reducing the variance and increasing classification performance. Nevertheless, enlarging the window length causes a delay in classification decision. The delay boundary of 300ms has been found to be acceptable and not perceived by the user [1], thus we could use windows of 100ms.

The dataset was then randomly splitted at 50% ratio into training and testing set. The first set was further reduced considering only the 10% for actual training. On the contrary, the first repetition of each movement from the latter was held out as validation set, while the remaining repetitions were used to form the testing set.

MAV and WL were chosen from the features named in chapter 2 to represent data; they were extracted separately for each channel.

SVM was used as classification method; the model was trained using the validation set to decide the best hyperparameters value, and finally tested on the testing set.

Moreover, a smoothing function was employed aiming to improve classification accuracy: it eliminates misinterpreted labels since it considers those with the greatest number of occurrences in a given sample, that is the most frequent.

The experiment sets were assembled and employed as follows.
Data were acquired during three sessions each day of the study, everyday at approximately the same time. Acquisitions were numbered from 2 to 14; a schematic view is given in Figure 10 to make it clear (the acquisition 1 was a trial acquisition, the 4th was missed because of an error in numbering).
The whole experiment was divided in two parts. The datasets were created as explained below (refer to Figure 11 for a schematic explanation).
The first acquisition of each day (respectively acquisitions 2, 6, 9, 12) was used as training set, gathering the 5% of the total samples. In the first part of the experiment the validation set was obtained combining data from the acquisitions of each session of the day, gathering the samples with the percentage defined above.
For instance, the training set used in the first experiment was obtained splitting the acquisition 2 dataset and considering the 10% of one half. The first repetition of each movement in the second half was held out to construct the validation set, jointly with the first movements repetition in the splitted dataset of acquisitions 3 and 5.
Data from every acquisition session in each day was tested. For all of them the testing set was composed by the repetitions of the movements left out from the validation set.
Training and testing sets in the second part of the experiment were the same as in the first part. By contrast, the validation set included a sample selection from the dataset of a single acquisition.

|           | 1st DAY | 2nd DAY | 3rd DAY | 4th DAY |
|-----------|---------|---------|---------|---------|
| session 1 | 2       | 6       | 9       | 12      |
| session 2 | 3       | 7       | 10      | 13      |
| session 3 | 5       | 8       | 11      | 14      |

Figure 10. Acquisition numbering, relating the day the data was acquired and the time of day. The three sessions took place respectively in the morning, earlier in the afternoon and in the evening.

|  | PART 1 ||||||||||||
|---|---|---|---|---|---|---|---|---|---|---|---|---|
| training set | 2 ||| 6 ||| 9 ||| 12 |||
| validation set | 2, 5, 6 ||| 6, 7, 8 ||| 9, 10, 11 ||| 12, 13, 14 |||
| testing set | 2 | 3 | 5 | 6 | 7 | 8 | 9 | 10 | 11 | 12 | 13 | 14 |

|  | PART 2 ||||||||||||
|---|---|---|---|---|---|---|---|---|---|---|---|---|
| training set | 2 ||| 6 ||| 9 ||| 12 |||
| validation set | 2 ||| 6 ||| 9 ||| 12 |||
| testing set | 2 | 3 | 5 | 6 | 7 | 8 | 9 | 10 | 11 | 12 | 13 | 14 |

Figure 11. Datasets composition in all the experiments performed.

## 5.2. Result

First of all, the model was trained using samples from the dataset of the first daily session (acquisitions 2, 6, 9 and 12), in order to select the best hyperparameters. For each day the experiments were performed attempting to classify data of the testing set. All the results are showed in Figures from A2 to A9 in Appendix; some examples are given below to highlight the most frequent behaviours.

The first experiment involved the model trained on acquisition 2 and tested on acquisition 2, 3 and 5 testing sets, using both MAV and WL separately for each case. Later, the smoothing function was applied. In Figure 12a the accuracy percentages are reported.

Figure 12b shows the accuracy values gained testing samples from acquisitions 6, 7 and 8 and training the model with the acquisition 6 training set.

Finally, Figure 12c gives an overview of the results obtained classifying data samples from acquisitions 9, 10 and 11; in these cases the acquisition 9 training set was employed.

In terms of classification accuracy no considerable improvement was observed representing data with different features, although WL generally gives an increment of the performance compared to MAV. Moreover, the smoothing function enhanced the accuracy values almost up to 3%.

In general, better results are given testing on the same acquisition the classifier was trained on and they decrease during the rest of the day.

Three different scenarios occurred: accuracy percentage keeps getting lower during the whole day (Figure 12a); it decreases in the middle of the day and then remains approximately constant (Figure 12b); it declines a lot in the middle acquisition and slightly increase in the last one (Figure 12c). In all of these cases, the performance reduction is significant only after the morning calibration; nevertheless the accuracy never decreases below 64%.

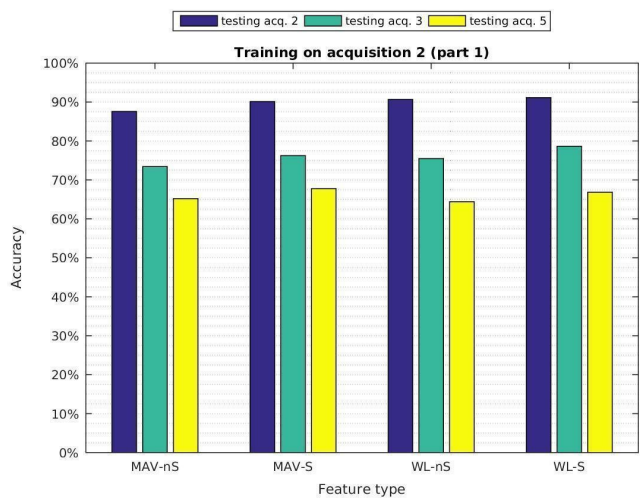 (a)
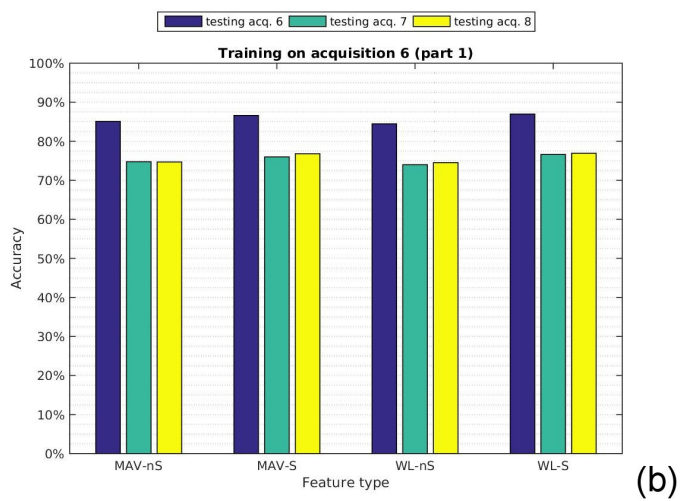 (b)
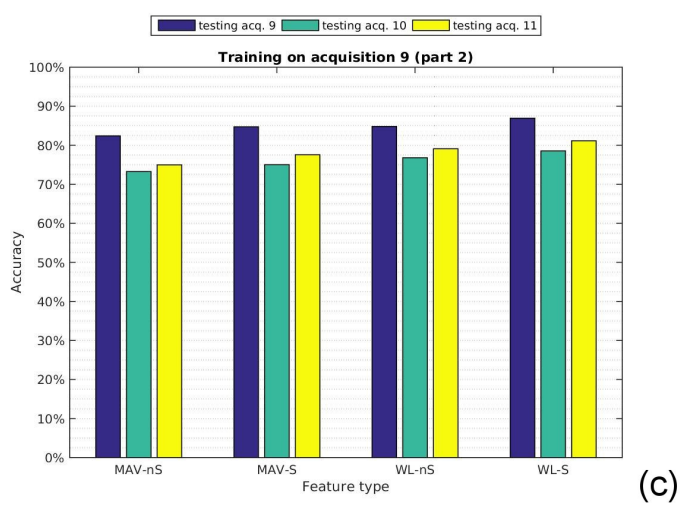 (c)

Figure 12. Accuracy values behaviour in three different day. Each group of histograms represents a specific feature used (MAV, WL), with (-S) or without (-nS) the smoothing function applied, and includes the acquisition at three different time.

Even if a larger number of samples in the training set ought to improve the model performance, no evidence has been found in this study. In fact, accuracy values in the first part of the experiment are not always higher than the one of the second part.

Regarding movement classification, visualizing accuracy values as a confusion matrix clearly shows a large number of correct predictions (diagonal components). However, the first column (corresponding to the rest position) indicates that almost every movement is misinterpreted, with varying percentages, as rest.

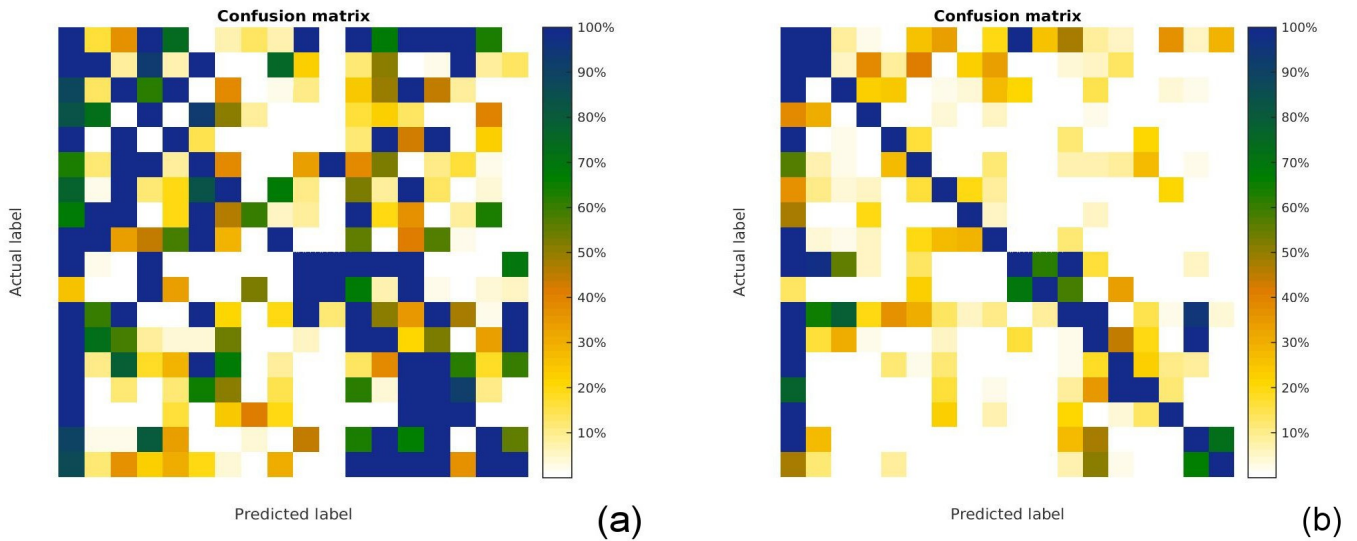

Figure 13. Confusion matrices respectively referred to (13a) the last acquisition of the first day (with data represented through MAV) and (13b) the first acquisition of the first day (with WL representation).

In particular, the confusion matrix of the worst case scenario (Figure 13a) appears having several high accuracy values corresponding to mispredicted movements. On the other hand, the best case scenario matrix (Figure 13b) shows that only few movements are misclassified. Furthermore, the first column presents less wrong movement predicted than usual.

It may be of interest that some movements seem to be more "learnable" than others, since they more often present an higher accuracy. For example, posture labeled as 3 in the Exercise 2 is usually well-predicted, thus it often appears among the first four movements (ordered by accuracy); in contrast, movement 9 has quite always the lowest accuracy value (Figure 14).

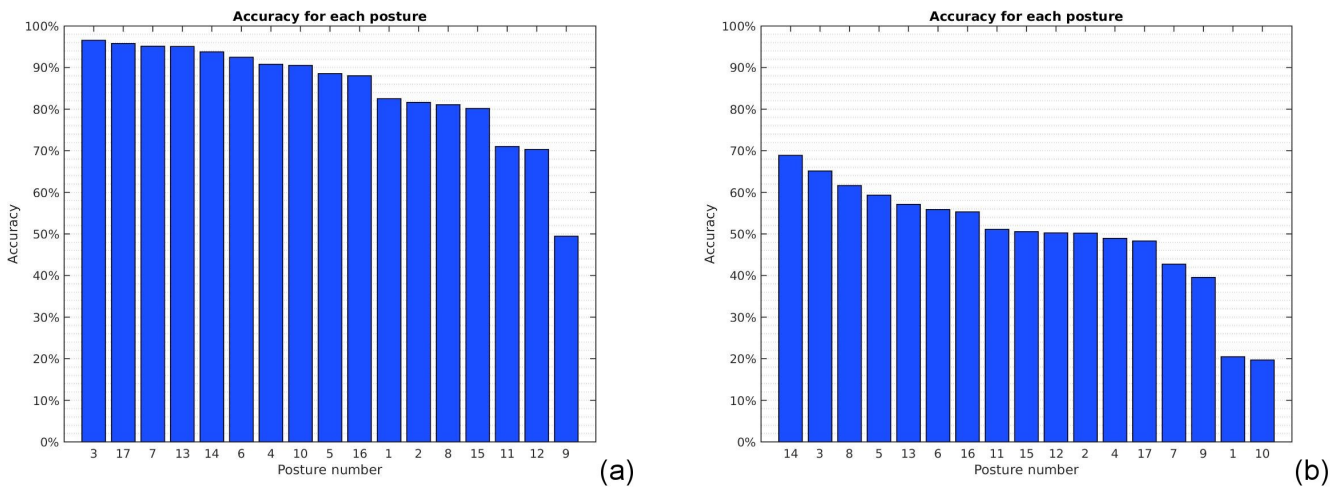

Figure 14. Movement labels ordered by accuracy value in the best (8a) and worst (8b) case scenario, using WL as feature.

## 5.3. Discussion

Naturally, the best accuracy overall is gained testing the model with the dataset of the same acquisition it was trained on.
Between the first and the second acquisition of the day it is always present a drop in performance (approximately 15%). This may be due to several factors; muscular fatigue and electrodes displacement, for example, are among them.

The different behavior at the end of the day cannot be related to any of the causes previously mentioned. Even the slight enhancement that sometimes occurs in the accuracy of the last daily acquisition is not considerable. In fact, it must be taken into consideration that a single subject was employed in this study, thus the results obtained cannot be reliable statistically speaking. Furthermore, the way the dataset is splitted may lead to different (though not dramatically so) performance result; therefore the accuracy increment at the end of the day might have not occurred changing the datasets (training, validation and testing) composition.

Reasonable causes of movements misclassification as rest might be several. For instance, windowing the sEMG signals, it may occurs that both rest and non-rest labels end up in the same window which is, thus, misleading.

# 6.  Conclusions

In the present work we carried out a performance evaluation of pattern recognition algorithm, focusing our study on the repeatability issue in a brief time period.
EMG signals were represented through two different features, Mean Absolute Value and Waveform Length, that appeared to have the best performance relating to their computational cost [6].
Support Vector Machine has been widely employed in previous works and it provided a satisfactory performance in our study, as well.

This work shows a practical equivalence between the features used to represent data (MAV and WL); in fact, although WL is slightly more efficient, the accuracy percentage differs for less than 3% in general.

The remarkable accuracy decrease that follows the first daily testing suggests a recalibration during the day might be needed. However, training the model may require a long time, and learning-based strategies might be required to do so with minimal discomfort for the user [12].
Anyway this result suggests the need to improve the device behavior especially in midday moments to get a general improvement, since the first daily acquisition tested always provides a better performance.

It appears to be usual that performance during the whole day (except for the first acquisition of the day) is almost constant and does not present noticeable decrease in accuracy. The general benchmark evaluation was satisfactory in the short period for every experiment performed except few rare cases. Those results might be caused by electrode misplacement during the day (since the whole setup has not been removed until evening), muscular fatigue or the particular random splitting of the dataset into training and testing set.

Only one subject was employed in this study and the dataset was acquired referred to 17 movements (adding rest). For these reasons a statistical analysis is not fully reliable, since it would require much more samples and experiments to be relevant. Hence future work ought to validate these results on a larger data collection.
Moreover, it might be interesting testing other classification algorithms to compare the result, combining with further feature types as well.

# Appendix

In this section, further graphs and table reporting gathered results are shown.

Accuracy percentage gained in all the experiments performed, with either composed and simple validation set, is displayed below. Both the features MAV and WL are reported. A comparison between the outcome before and after the smoothing function application is provided.

|  | Training set | Validation set | Testing set | Accuracy - MAV | | Accuracy - WL | |
|---|---|---|---|---|---|---|---|
|  |  |  |  | no Smoothing | with Smoothing | no Smoothing | with Smoothing |
| PART 1 | 2 | 2,3,5 | 2 | 87,58 | 90,13 | 90,65 | 91,13 |
|  |  |  | 3 | 73,43 | 76,24 | 75,5 | 78,65 |
|  |  |  | 5 | 65,19 | 67,77 | 64,4 | 66,85 |
|  | 6 | 6,7,8 | 6 | 85,08 | 86,6 | 84,45 | 86,96 |
|  |  |  | 7 | 74,75 | 75,99 | 73,99 | 76,62 |
|  |  |  | 8 | 74,68 | 76,8 | 74,52 | 76,92 |
|  | 9 | 9,10,11 | 9 | 83,7 | 84,32 | 85,22 | 86,76 |
|  |  |  | 10 | 75,03 | 75,97 | 77 | 79,57 |
|  |  |  | 11 | 78,08 | 78,3 | 79,95 | 82,31 |
|  | 12 | 12,13,14 | 12 | 85,42 | 85,86 | 86,93 | 87,39 |
|  |  |  | 13 | 73,66 | 75,13 | 73,02 | 74,14 |
|  |  |  | 14 | 75,25 | 76,47 | 74,57 | 76,67 |
| PART 2 | 2 | 2 | 2 | 86,92 | 89,26 | 90,56 | 91,13 |
|  |  |  | 3 | 74,47 | 76,75 | 75,5 | 78,65 |
|  |  |  | 5 | 65,06 | 67,14 | 64,4 | 66,85 |
|  | 6 | 6 | 6 | 85,44 | 86,83 | 84,54 | 86,85 |
|  |  |  | 7 | 74,1 | 75,54 | 74,26 | 76,66 |
|  |  |  | 8 | 73,5 | 75,83 | 74,97 | 77,7 |
|  | 9 | 9 | 9 | 82,39 | 84,72 | 84,82 | 86,93 |
|  |  |  | 10 | 73,26 | 75,03 | 76,81 | 78,55 |
|  |  |  | 11 | 74,98 | 77,58 | 79,1 | 81,13 |
|  | 12 | 12 | 12 | 86,29 | 86,75 | 85,88 | 86,06 |
|  |  |  | 13 | 72,24 | 73,01 | 74,5 | 75,63 |
|  |  |  | 14 | 73,36 | 74,6 | 75,64 | 77,52 |

The following histograms represent the accuracy comparison in the three daily sessions. Each graph gathers the results of a single day. On the left column are showed results from the first part of the experiment, on the right one from the second part (description of the experiments in chapter 5.1).
For each image the acquisition used as training set is specified in the title; the legend lists the acquisition datasets that were employed as testing set in the experiments.

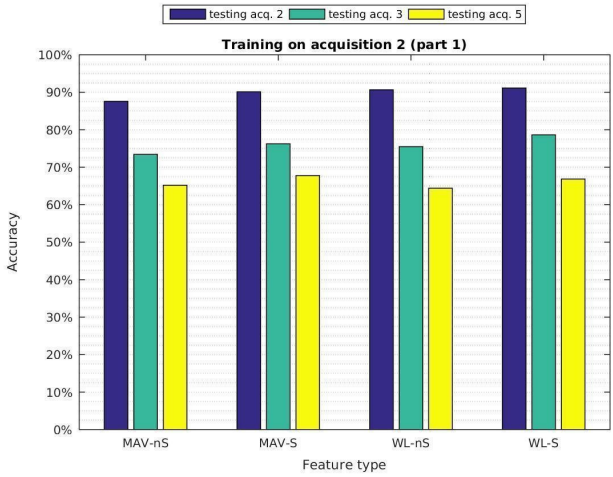
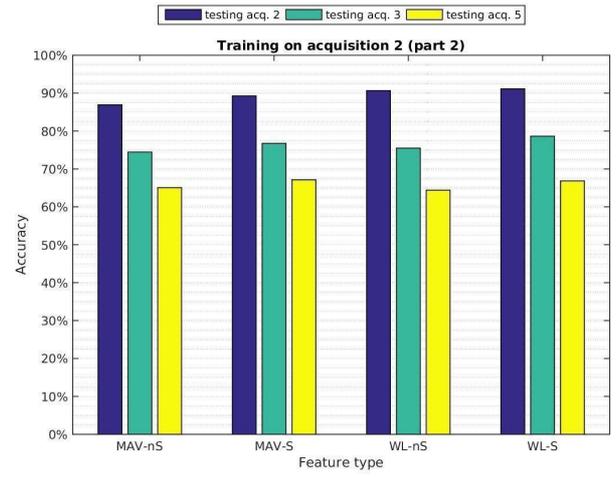
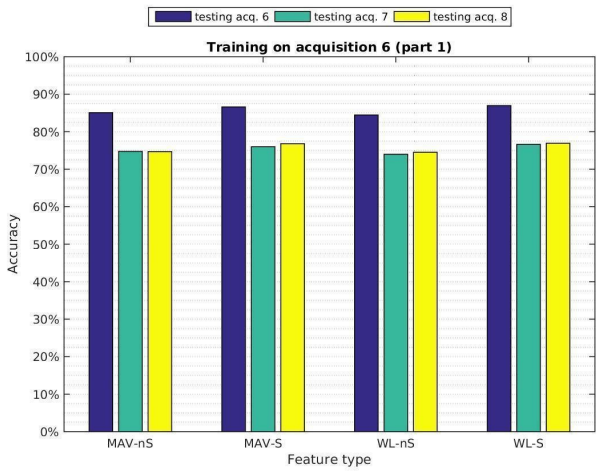
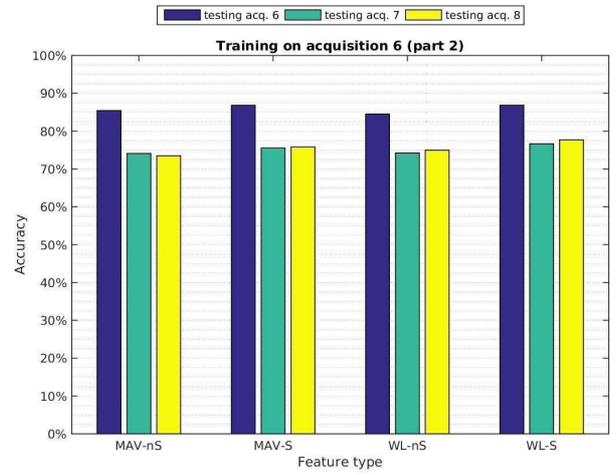
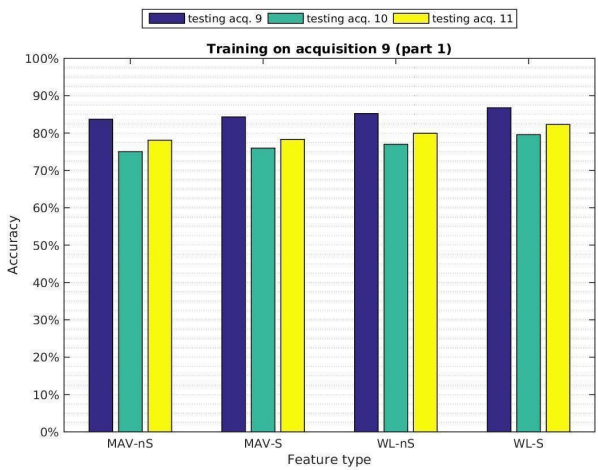
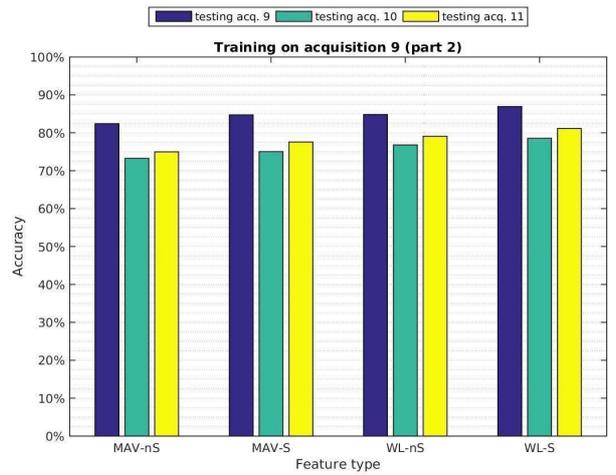

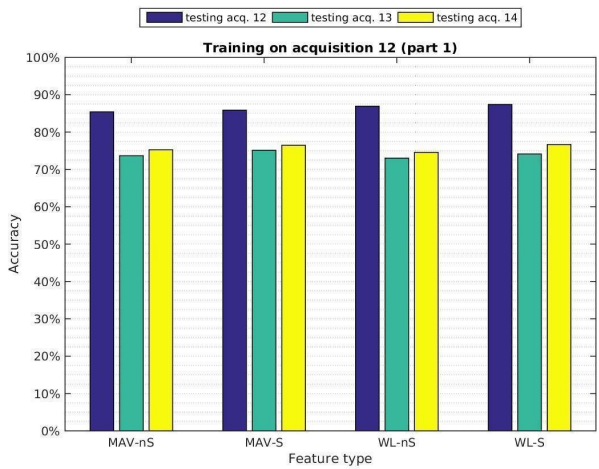
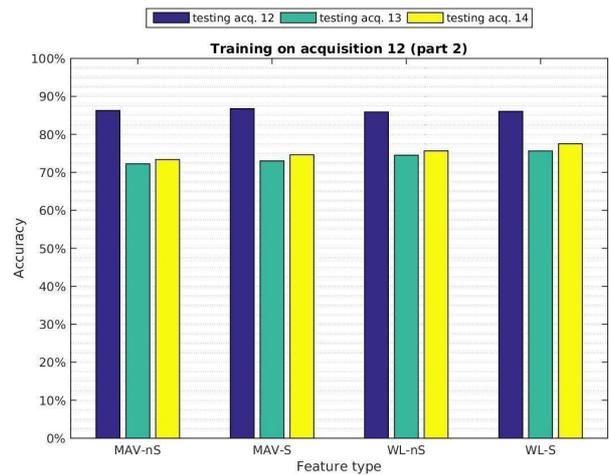

The following table presents a comparison between the best, worst and average accuracy overall, in the best case scenario and in the worst case scenario. In particular, either MAV and WL performance is displayed, before and after the smoothing function.

Among all the experiments, the best case in each day appeared to be the acquisition gained during the first daily session. The last acquisition of the day, by contrast, provided the worst accuracy in every case, referring to a particular day.

|  | Overall | | | Best Case Scenario | | | Worst Case Scenario | | |
|---|---|---|---|---|---|---|---|---|---|
|  | best | worst | average | best | worst | average | best | worst | average |
| MAV - no smoothing | 87,58 | 65,06 | 77,24 | 87,58 | 82,39 | 85,35 | 78,08 | 65,06 | 72,51 |
| MAV - smoothing | 90,13 | 67,14 | 78,86 | 90,13 | 84,32 | 86,81 | 78,3 | 67,14 | 74,31 |
| WL - no smoothing | 90,65 | 64,4 | 78,39 | 90,65 | 84,45 | 86,64 | 79,95 | 64,4 | 73,44 |
| WL - smoothing | 91,13 | 66,85 | 80,32 | 91,13 | 86,06 | 89,9 | 82,31 | 66,85 | 75,74 |

The outcomes in the previous table are here represented through histograms. The best accuracy is displayed on the left, the worst on the right. Graphs in the three row are respectively referred to the whole experiment, the best case and the worst case.

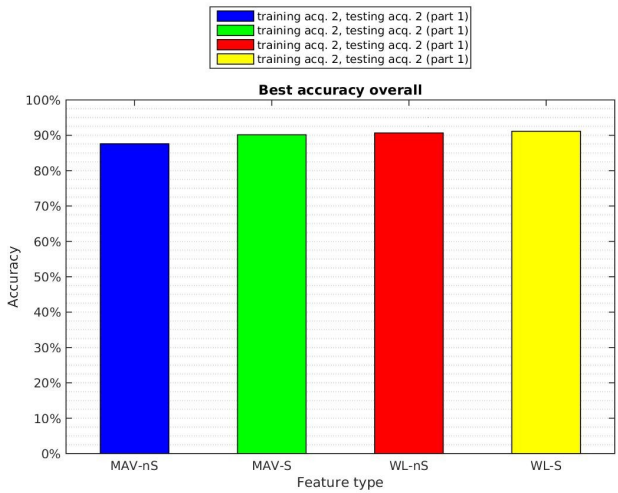
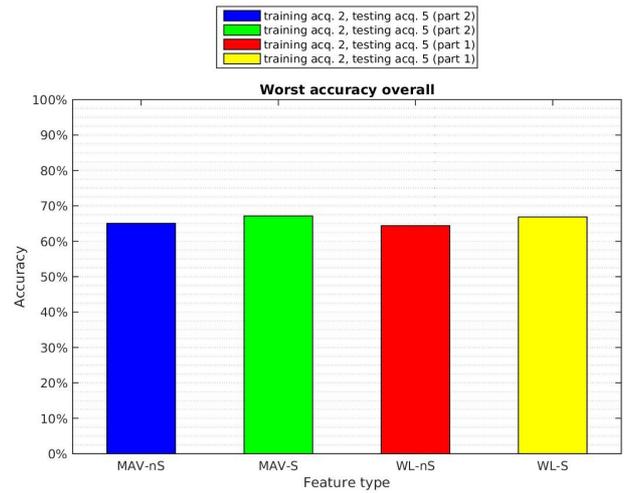
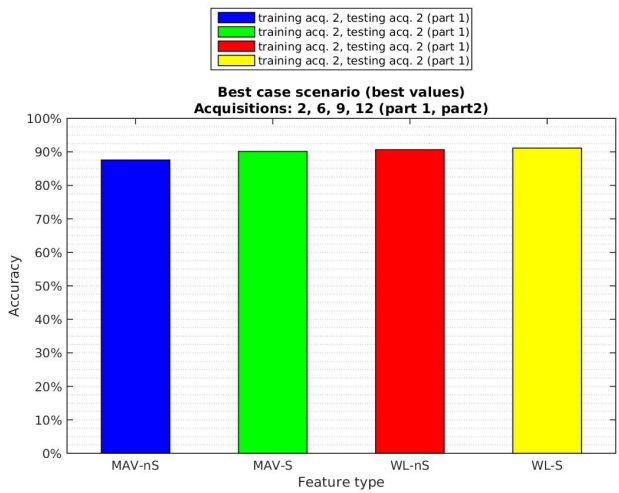
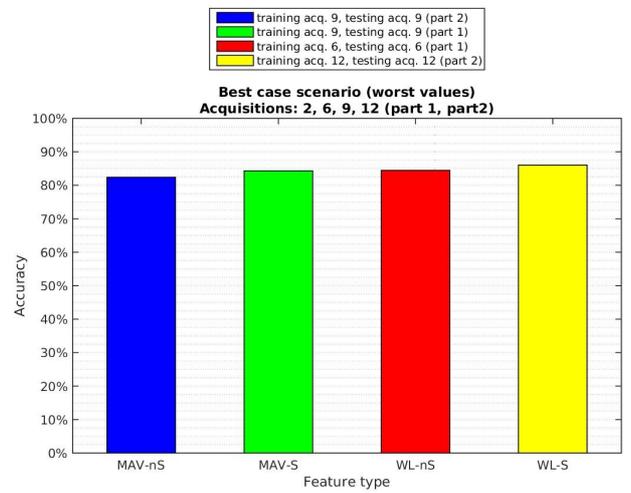
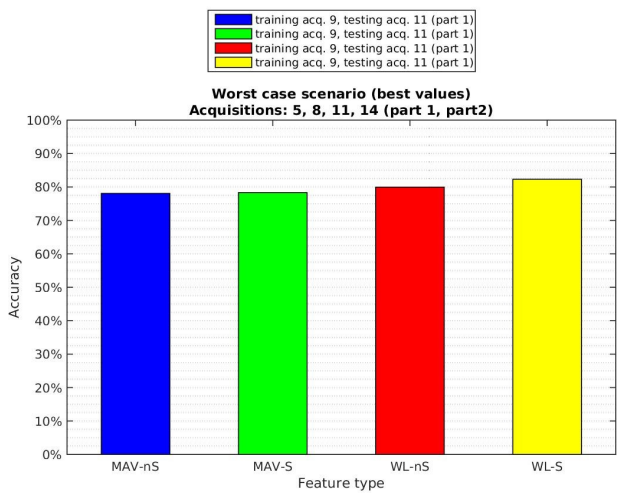
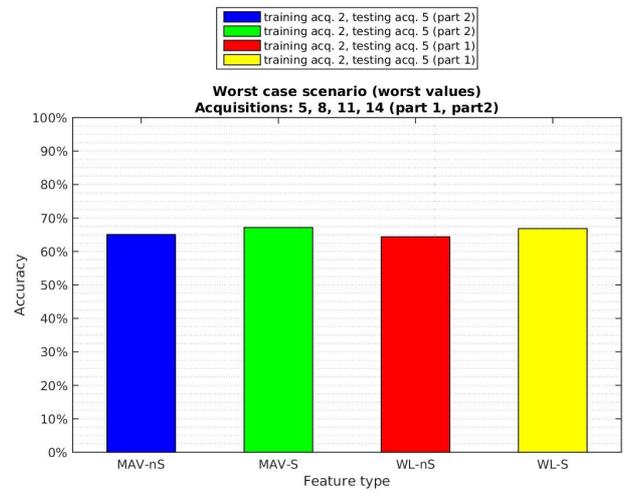

The average accuracy in all the three cases is showed below.

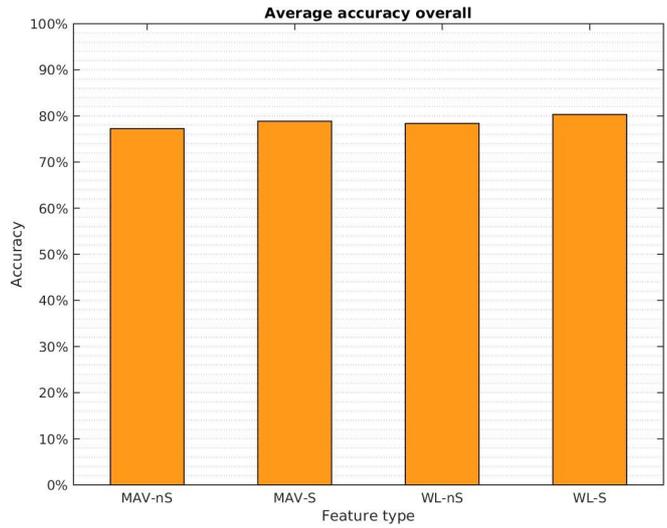

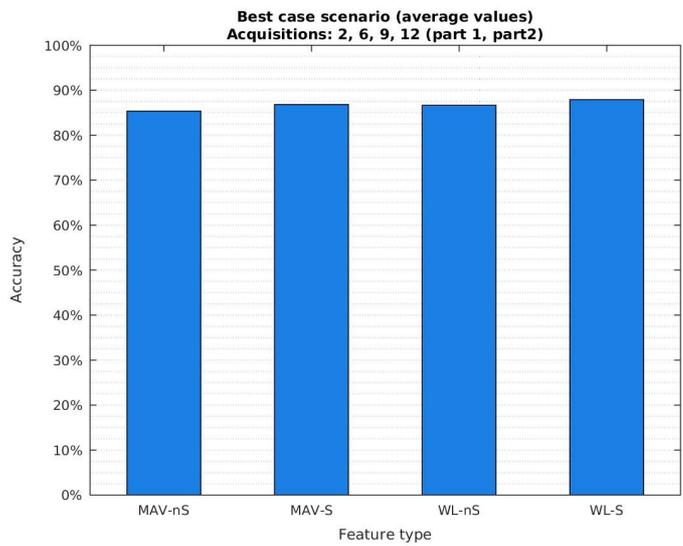

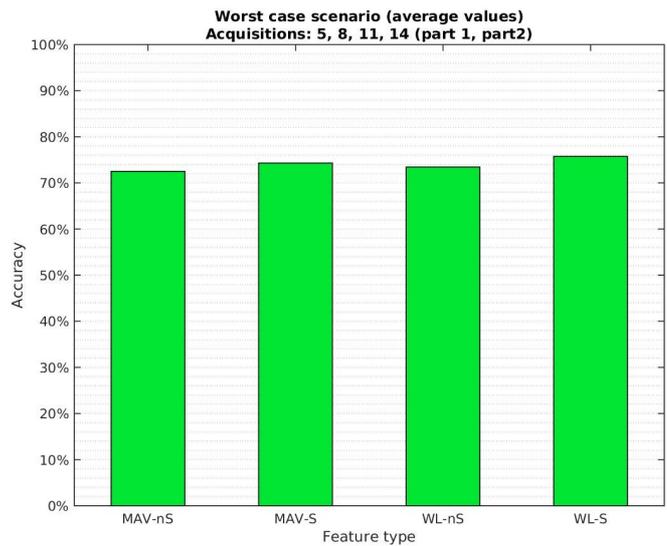

In this last image, the best, worst and average accuracy referred to each movement label are shown as histograms, in order to have a complete comparison between all of them.

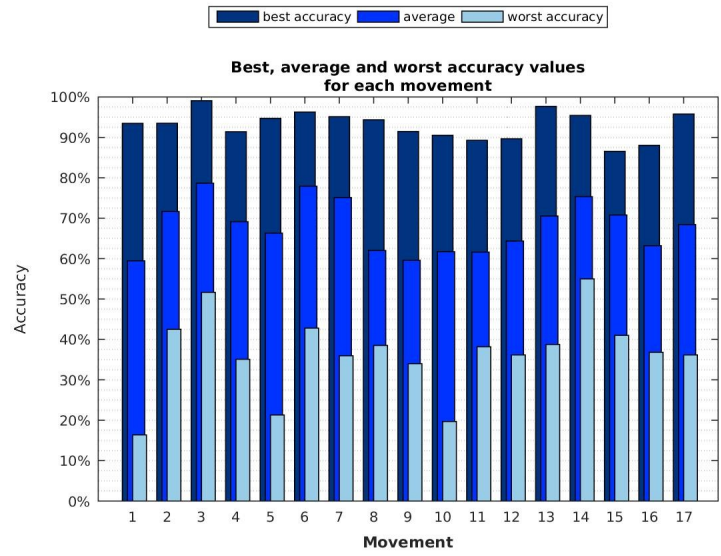